\documentclass[acmsmall]{acmart}

\usepackage{subfig}
\newlength{\myheightfirst}
\newlength{\myheightsecond}

\AtBeginDocument{%
  }

\setcopyright{cc}
\setcctype{by}
\acmJournal{PACMCGIT}
\acmYear{2025} \acmVolume{8} \acmNumber{2} \acmArticle{21} \acmMonth{6} \acmPrice{}\acmDOI{10.1145/3729413}

\begin{document}

\title[Iris Style Transfer]{Iris Style Transfer: Enhancing Iris Recognition with Style Features and Privacy Preservation through Neural Style Transfer}

\author{Mengdi Wang}
\orcid{0000-0002-5254-737X}
\affiliation{%
  \institution{Technical University of Munich}
  \city{Munich}
  \country{Germany}
}
\email{mengdi.wang@tum.de}

\author{Efe Bozkir}
\orcid{0000-0002-4594-4318}
\affiliation{%
  \institution{Technical University of Munich}
  \city{Munich}
  \country{Germany}
}
\email{efe.bozkir@tum.de}

\author{Enkelejda Kasneci}
\orcid{0000-0003-3146-4484}
\affiliation{%
  \institution{Technical University of Munich}
  \city{Munich}
  \country{Germany}
}
\email{enkelejda.kasneci@tum.de}

\renewcommand{\shortauthors}{Wang, et al.}

\begin{abstract}
Iris texture is widely regarded as a gold standard biometric modality for authentication and identification. The demand for robust iris recognition methods, coupled with growing security and privacy concerns regarding iris attacks, has escalated recently. Inspired by neural style transfer, an advanced technique that leverages neural networks to separate content and style features, we hypothesize that iris texture's style features provide a reliable foundation for recognition and are more resilient to variations like rotation and perspective shifts than traditional approaches. Our experimental results support this hypothesis, showing a significantly higher classification accuracy compared to conventional features. Further, we propose using neural style transfer to obfuscate the identifiable iris style features, ensuring the protection of sensitive biometric information while maintaining the utility of eye images for tasks like eye segmentation and gaze estimation. This work opens new avenues for iris-oriented, secure, and privacy-aware biometric systems.
\end{abstract}



\received{1 November 2024}
\received[revised]{18 February 2025}
\received[accepted]{13 March 2025}

\maketitle

\section{Introduction}
Reliable and robust automatic recognition of users using biometrics~\cite{jain2004introduction, jain2007handbook} has been longed for decades. The iris, a rich source of unique biometric information, features distinctive textures and patterns that can reliably differentiate individuals. Due to its identifiability, uniqueness, and favorable mathematical properties~\cite{flom1987iris, daugman2003importance, daugman2009iris, jain2004introduction, ma2004efficient}, the iris has come to be regarded as a gold standard in biometrics. This has led to the widespread adoption of iris-based identification and authentication methods across a variety of devices, from dedicated iris scanners to VR/AR/MR (virtual/augmented/mixed reality) headsets and smart glasses, where reliability and convenience are paramount. Notably, Apple’s Vision Pro integrates Optic ID for iris-based authentication~\cite{visionpro}, and Microsoft's HoloLens 2 also supports iris login on top of Windows Hello~\cite{hololens}. While most head-mounted displays currently lack built-in iris recognition functionality, devices such as Meta Quest Pro and Magic Leap 2, which include eye-tracking capabilities, have promising potential to enable software-based iris recognition, as eye-tracking and iris recognition techniques share several underlying principles. 

Despite the expanding use of iris recognition, the impact of factors like image quality, sensor setup, and feature robustness under these varying conditions remains under-explored. At the same time, the popularity of iris recognition has sparked a rise in security threats like presentation attacks~\cite{sequeira2016realistic, kohli2017synthetic, narkar2024swap}. Although countermeasures have been developed~\cite{sequeira2014mobilive, sequeira2014iris, doyle2015robust, rigas2015eye, raja2015video, czajka2018presentation, raju2022iris}, concerns about the balance between privacy, security, and utility in iris-based systems continue to grow. Striking a sustainable trade-off between protection and usability~\cite{john2020security} is now more critical than ever.

Neural style transfer~\cite{gatys2015neural, gatys2016image} opens up a fresh opportunity for tasks like image synthesis and artistic blending. It operates on the premise that each image is a combination of content and style. By leveraging convolutional neural networks (CNN), which are commonly trained for image classification tasks, neural style transfer strives to separate content and style features and then recombine them in a unique way. It provides a novel approach to interpret style features of an image statistically by capturing correlations between feature activations and can be hence regarded as a special domain adaptation problem~\cite{li2017demystifying}.

As the style features of different images extracted by neural style transfer show quite distinctive properties, in this paper, we propose using iris style features for iris recognition. Considering that style features are extracted in the form of global statistical distribution, we hypothesize that iris style features, represented as global statistical distributions, are inherently robust to image variations such as rotation and perspective transformation due to their non-dependence on individual pixel positioning. In this regard, we investigate the impact of random rotation and perspective transformation of different degrees on iris recognition using traditional CNN features and style features on the OpenEDS2019~\cite{garbin2019openeds} dataset. Our results show that style features deliver significantly better test classification accuracy in both cases. Additionally, we suggest using neural style transfer to erase the sensitive iris style features to obscure the biometric identities of users without ruining image quality and the utility of eye images for downstream tasks like eye semantic segmentation and gaze estimation. We validate the effectiveness of our privacy-preservation method by presenting a prominent drop in recognition accuracy and a little to no decrease in segmentation accuracy on the OpenEDS2019 dataset, as well as a neglectable increase in gaze estimation error on the OpenEDS2020 dataset~\cite{palmero2020openeds2020, palmero2021openeds2020}. The main contributions of this work are four-fold:
\begin{itemize}
    \item We propose a novel feature extraction method for the iris feature using neural style transfer, and suggest the use of the iris style feature for iris recognition accordingly.
    \item We investigate the impact of random image rotation and perspective transformation of different transformation degrees, which are common problems with cameras as input modality, on iris recognition using conventional CNN features and style features.
    \item We show that iris style features are remarkably more robust against random rotation and perspective transformation than traditional CNN features. Our results lay the foundation of iris recognition utilizing iris style features.
    \item We suggest the use of neural style transfer to blend out the sensitive iris style features in eye images to conceal user identity from imposters without sacrificing much the quality and the utility of eye images. We empirically validate the feasibility of our approach with a significant drop in recognition accuracy and negligible to zero changes in image quality, eye segmentation accuracy, and gaze estimation error. Besides, we show that the risk of malicious use of our work by adversaries is relatively low.
\end{itemize}

To the best of our knowledge, this is the first work to present neural style transfer for iris recognition and privacy preservation. The proposed iris style transfer prototype can be easily extended from static eye images to dynamic eye videos in real time on modern devices. Our work addresses a new research gap in this regard. We also publish
our code\footnote{~\url{https://gitlab.lrz.de/hctl/Iris-Style-Transfer}.\label{footnote:url}} for reproducibility.

\section{Related Works}

\subsection{Iris Recognition}
As iris contains unique biometric features and demonstrates distinct mathematical advantages~\cite{flom1987iris, daugman2003importance, daugman2009iris, jain2004introduction, ma2004efficient}, it has been researched as user identification and authentication modality for long. The idea of iris recognition can be traced back to the early days when iris features were usually encoded by handcrafted filters and kernels~\cite{daugman2009iris, wildes1997iris, ali2007recognition, hong2005iris} and dependent on meticulously chosen metrics like Hamming Distance. Given the giant stride of deep learning, a shift towards neural network-based feature extraction for iris patterns has been experienced.~\cite{liam2002iris} used a single layer network to extract iris features. A growing body of research demonstrates the benefits of convolutional neural networks as feature extractors. For instance,~\cite{boyd2019deep} benchmarked five different features encoded by ResNet50~\cite{he2016deep} using different weights.~\cite{minaee2016experimental} employed a VGG16~\cite{simonyan2014very} as representation encoder. In~\cite{hafner2021deep}, the authors integrated a DenseNet201~\cite{huang2017densely} into traditional pipeline as an iris embedding generator. For a broader view in this regard, we refer readers to the survey works~\cite{yin2023deep, nguyen2024deep}.

Despite the strong momentum of iris-based authentication and identification, the influence of factors like image quality, resolution, and perspective distortions are frequently under-explored, particularly within neural network-based methods.~\cite{wei2005robust, kalka2006image, galbally2013image, wang2020recognition} generally discussed image quality evaluation criteria from different perspectives.~\cite{phillips2011impact} specialized in their exploration of the impact of image resolution and blurring on iris identification.~\cite{ribeiro2017exploring} investigated the effect of image super-resolution. While traditional methods account for rotation to some extent ~\cite{daugman2009iris, daugman2003importance}, its impact on neural network-rooted embeddings remains unclear. An even less discussed factor is perspective transformation, which introduces viewpoint distortions to images based on sensor positioning and is generally more challenging to capture than affine transformation in deep learning. This work uniquely evaluates style features' robustness under these challenging conditions, showing potential for greater resilience than traditional CNN-based embeddings.

\subsection{Iris Privacy and Attack}
As eye presentation is ubiquitous and people often tend to be less conscious about iris privacy compared to other biometrics like facial data and fingerprints~\cite{liebling2014privacy}, the leak of iris information can be escalated, which could lead to undesired iris attacks given the surge of iris recognition. One of those most common attacks is the iris presentation attack, which describes the case that an imposter impersonates someone's identity by presenting a fake iris sample of the victim to the camera, either printed on paper or monitor or even contact lenses. Although many works have been contributed to triggering or detecting iris presentation attacks~\cite{sequeira2014mobilive, sequeira2014iris, rigas2015eye, doyle2015robust, raja2015video, sequeira2016realistic, kohli2017synthetic, czajka2018presentation, raju2022iris}, the defense against iris presentation attacks are still difficult as the attacks occur beyond the physical boundaries of the system~\cite{nguyen2024deep}. Until now, only a few works have strived to tackle the challenge on the image source side.~\cite{john2019eyeveil} proposed to optically defocus iris image while maintaining eye tracking utility.~\cite{john2020let} imputed noise following differential privacy into eye images to safeguard user identity. While in~\cite{chaudhary2020privacy} the authors replaced the raw iris texture with fake templates to protect user privacy, their work can be applied in the opposite direction by swapping in a victim's iris texture to fake his presentation~\cite{narkar2024swap}.

\subsection{Neural Style Transfer}
\label{sec:nst}
\begin{figure}[h]
  \centering
  \includegraphics[width=10cm]{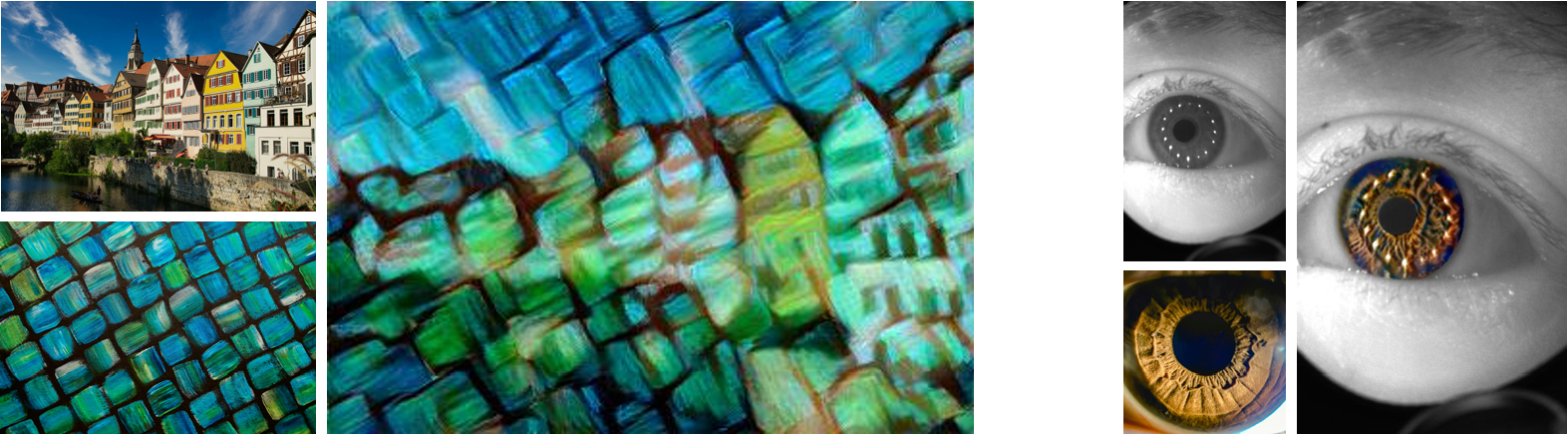} 
  \caption{Two examples of neural style transfer. Eye image sources from OpenEDS2019~\cite{garbin2019openeds}.}
  \label{fig:nst_example}
  \Description{Two examples of neural style transfer.}
\end{figure}

Neural style transfer~\cite{gatys2015neural, gatys2016image} is a promising image synthesis technique that can be categorized as a branch of non-photorealistic rendering~\cite{gooch2001non, strothotte2002non}. The fundamental assumption of neural style transfer is that the content and style of an image can be separated and manipulated independently. While serving different purposes, a similar idea of eye image decomposition has been presented in~\cite{john2020security}. By disentangling the content and style components and recombining the style parts of other images, neural style transfer can achieve amazing image stylization effects. Two examples of neural style transfer are given in Figure~\ref{fig:nst_example}. In its original form, neural style transfer extracts content and style features using a VGG~\cite{simonyan2014very} network, which is pre-trained on the ImageNet~\cite{deng2009imagenet} dataset for classification purposes. While the content features are, in essence, activations from convolutional layers, the style features are originally computed as correlations of flattened convolutional embeddings using a gram matrix. Then, neural style transfer operates by iteratively optimizing pixel values of the output image for content and style feature matching objectives. In~\cite{li2017demystifying}, the authors proved that the matching of gram matrices is equivalent to minimizing the Maximum Mean Discrepancy~\cite{gretton2012kernel} with a polynomial kernel of second order. Hence, neural style transfer can be regarded as a domain adaptation problem in the CNN layer space. In~\cite{wright2022artfid}, the researchers showed that the style features have strong uniqueness and identifiability.

In its vanilla form, neural style transfer is an image-optimization-based online algorithm~\cite{jing2019neural} and functions slowly due to its iterative nature. Many follow-up works~\cite{johnson2016perceptual, gupta2017characterizing, huang2017arbitrary} reform the style transfer as a model-optimization-based offline procedure, meaning that after training a stylization model, style transfer can be approached with a simple and fast feed-forward through the network. While the ordinary style transfer is only designed for static images, it can be applied to videos by processing frame by frame, which often leads to flickering because of inconsistent initialization and stochastic optimization. In recent years many works have been contributed to accommodate neural style transfer for videos, even realizing video stylization in real time~\cite{ruder2016artistic, chen2017coherent, huang2017real, gao2019reconet, liu2021adaattn}.

While substantial progress has been made in both iris recognition and privacy-preserving techniques, existing methods primarily focus on traditional feature extraction and privacy frameworks that may not fully exploit the unique properties of iris style. Neural style transfer, with its capacity to extract robust style representations, offers an untapped potential for enhancing iris recognition and safeguarding privacy in biometric systems. Building on this potential, our work introduces a novel framework leveraging iris style features to address both robustness and privacy concerns in a unified approach.

\section{Methodology}
We observe that iris patterns contain special brushes and strokes that resemble artistic works to some extent. As style features are highly distinctive~\cite{wright2022artfid} and play the core role of distribution alignment~\cite{li2017demystifying} for neural style transfer, we claim that the iris style feature extracted by neural style transfer has a good potential to serve as a biometric identifier and offers better resilience to random image rotation and perspective transformation by its nature. Further, we propose to transfer iris style to erase the sensitive iris style features for privacy protection purposes. In the following, we first describe the iris style feature extraction pipeline in Section~\ref{sec:feature_extraction}, and then explain the iris style transfer procedure in Section~\ref{sec:iris_nst}.

\subsection{Feature Extraction}
\label{sec:feature_extraction}
\begin{figure}[!ht]
  \centering
  \includegraphics[width=13.5cm]{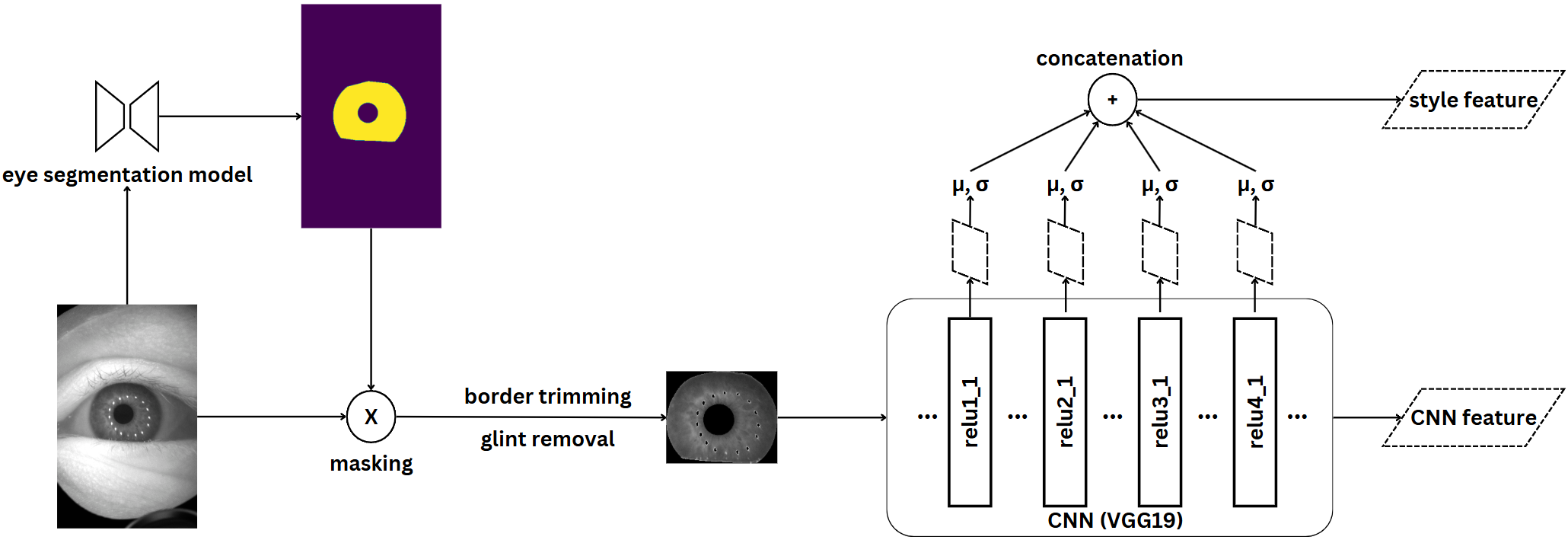} 
  \caption{The iris style feature extraction pipeline. Eye image sources from OpenEDS2019~\cite{garbin2019openeds}.}
  \label{fig:pipeline}
  \Description{The iris style feature extraction pipeline.}
\end{figure}
Our iris style feature extraction pipeline is depicted in Figure~\ref{fig:pipeline}. First of all, the eye image is fed into an eye region segmentation model to mask out the non-iris region. Then, we trim the black border of the eye image. Further, we filter out the glints (reflection of sensor light) by applying a threshold. In the next step, the iris image passes through a CNN, and the activations computed at some specific layers of the CNN will be utilized for style feature calculation. For each representation extracting layer in the network, the vanilla neural style transfer self-correlates its activation using a gram matrix, which results in a feature map of complexity $O(N^2)$ with $N$ being the number of embedding channels.~\cite{li2017demystifying} suggested the use of channel-wise batch normalization mean and standard deviation ~\cite{ioffe2015batch} as style representations, which contain domain traits~\cite{li2016revisiting} and tend to be more stable in neural style transfer compared to correlation-based feature set~\cite{wang2022exploration}. Therefore, we considered the normalization statistics as our iris style feature modality. A further benefit of the statistics feature is that the feature complexity reduces from $O(N^2)$ to $O(2N)$. In detail, given the activation map $F^l \in \mathbb{R}^{N^l \times H^l \times W^l}$ of layer $l$, the style features at this layer are computed as the following channel-wise statistics:
\begin{equation}
    \mu^i_{F^l} = \frac{1}{H^l W^l} \sum_{j=1}^{H^l} \sum_{k=1}^{W^l} F^l_{i,j,k}\:, \quad {\sigma^i_{F^l}}^2 = \frac{1}{H^l W^l} \sum_{j=1}^{H^l} \sum_{k=1}^{W^l} (F^l_{i,j,k} - \mu^i_{F^l})^2
\end{equation}
where $i \in [N^l]$ indices to feature map channel. 

Considering that the pipeline encodes style features in the form of global statistical distribution parameters regardless of individual pixel and activation positioning in the feature map, we believe that the iris style feature is more robust against image variations in contrast to the common CNN feed-forward features by its nature. While image quality problems like blurring and low resolution can be mostly alleviated with close-eye cameras, variations like rotation and perspective transformation can still be problematic and inevitable for camera-based sensing systems. We conduct extensive experiments to evaluate the robustness of the iris style feature and the common CNN feature in Section~\ref{sec:robustness}. Our results show that the iris style feature offers notably more resilience to random rotation and perspective transformation concerning iris recognition performance.

\subsection{Iris Style Transfer}
\label{sec:iris_nst}
Since iris style feature proves to be valuable for iris recognition, we move our concerns towards iris privacy preservation by swapping the raw iris style in images. Our goal is to obscure iris-based authentication and identification using stylized eye images without sacrificing the quality, realism, and utility of the images, such as being used for eye segmentation and gaze estimation purposes. We approach our goal with the help of neural style transfer. By transferring the style of another image (not necessarily an eye image) to the iris texture, the original iris style feature is blended out and cannot be used for iris recognition anymore. We term this procedure as iris style transfer. 

Given an eye image $I_C$ that shall be stylized and a style-providing image $I_S$, the output image $I_X$ is normally initialized as a clone of $I_C$. Assume a layer $l$ is chosen for feature extraction and objective computation. All three images are then fed into the feature extraction pipeline mentioned above to extract the feature maps $F^l_C, F^l_S, F^l_X \in \mathbb{R}^{N^l \times H^l \times W^l}$ and the corresponding batch normalization statistics $\mu_{F^l_C}, \sigma_{F^l_C}, \mu_{F^l_S}, \sigma_{F^l_S}, \mu_{F^l_X}, \sigma_{F^l_X} \in \mathbb{R}^{N^l}$ at layer $l$. The style transfer operates through iteratively optimizing the output image regarding a weighted combination of content and style losses:
\begin{equation}
    \mathcal{L}^l_{total} = \alpha \mathcal{L}^l_{content} + \beta \mathcal{L}^l_{style}
\end{equation}
where the content objective is computed as the mean squared error between the features maps:
\begin{equation}
    \mathcal{L}^l_{content} = \frac{1}{N^l H^l W^l} \sum_{i=1}^{N^l} \sum_{j=1}^{H^l} \sum_{k=1}^{W^l} (F^l_{C_{i,j,k}} - F^l_{X_{i,j,k}})^2 
\end{equation}
and the style loss is calculated by aligning distribution means and standard deviations:
\begin{equation}
    \mathcal{L}^l_{style} = \frac{1}{N^l} \sum_{i=1}^{N^l} \left( \left(\mu^i_{F^l_S} - \mu^i_{F^l_X} \right)^2 + \left(\sigma^i_{F^l_S} - \sigma^i_{F^l_X}\right)^2 \right)
\end{equation}
It should be noted that in practice, the content loss is often calculated merely at a single layer, while the computation of style loss normally involves multiple layers that are weighted differently. 

After obtaining the stylized iris texture, it is used to replace the original iris part with the help of an eye segmentation mask. The glints shall also be added back at this time, as many eye-tracking algorithms rely on the reflections. In our experiments, for each eye image, we randomly sample an eye image from another user and transfer its style. The impact of iris style transfer on iris recognition accuracy is analyzed in Section~\ref{sec:privacy}.

\section{Experiment}
We conducted our experiments using two datasets, namely OpenEDS2019~\cite{garbin2019openeds} and OpenEDS2020~\cite{palmero2020openeds2020, palmero2021openeds2020}. All experiments were executed on a cluster equipped with AMD EPYC 7763 64-Core processor and 4$\times$ NIVIDA A100 80GB graphic cards. An overview of our environment is given in Table~\ref{tab:appendix_environment} in the appendix. Our code is published$^{\ref{footnote:url}}$ for reproducibility.

\subsection{Datasets, Tasks, and Models}
\subsubsection{Datasets} \label{datasets}
\sloppy
We employed two datasets, namely OpenEDS2019~\cite{garbin2019openeds} and OpenEDS2020~\cite{palmero2020openeds2020, palmero2021openeds2020}, in our experiments. The OpenEDS2019 dataset is tailored for eye region segmentation. It was collected from 152 users using a VR headset with two close-eye cameras, containing 12,759 eye images of both eyes with segmentation annotations. On the contrary, its successor, OpenEDS2020, is a dataset collected in a similar environment while mainly specialized for gaze estimation and prediction. It consists of 1,280 100-frame sequences in the training set and 7,680 55-frame sequences in the validation and test sets. All samples in both datasets are 8-bit grayscale images, i.e., each pixel is a single value ranging from 0 to 255. An overview of the datasets is given in Table~\ref{tab:appendix_datasets}. It should be noticed that although the two datasets were collected in similar environments, their setups were still different in many aspects. Particularly, the image resolution of OpenEDS2019 is $400 \times 640$, whereas the image shape of OpenEDS2020 is $640 \times 400$. Therefore, models that are trained on one dataset often do not work seamlessly on the other one. For OpenEDS2019, since the original data split was done user-wise, which is not suitable for user recognition, we conducted a sample-wise data split by ourselves. We left out users who have less than 2 samples, resulting in 151 distinct classes. We utilized 20\% samples per user for testing. For OpenEDS2020, we kept the original data split. Although both datasets contain images from both left and right eyes, labels for indicating left and right eyes are missing; hence, we did not design benchmarks regarding ocular laterality.

\subsubsection{Tasks} \label{sec:tasks}
Our experiments can be divided into two task groups. In the first group, we investigated the feasibility of iris style feature-driven user recognition and explored the robustness of iris style features against random image rotation and perspective transformation. Throughout our experiments in the first task group, we emulated iris recognition with an iris-based classification task, meaning that each user in the dataset forms a unique class. In the second group, we shifted our attention to the impact of iris style transfer, including its influence on image quality, privacy, data utility, and the risk of malicious use. Particularly, to investigate the impact of our pipeline on data utility, we compared two utility tasks, namely eye segmentation and gaze estimation, before and after conducting iris style transfer. To study the risk of malicious use, for each eye image, we randomly transferred the iris style of another user and focused on the false acceptance rate (FAR). The first task group, as well as the experiments regarding privacy, eye segmentation, and risk of malicious use, were carried out on the OpenEDS2019 dataset, while the impact on gaze estimation was studied on the OpenEDS2020 dataset.

\subsubsection{Models} \label{sec:models}
The iris style-based recognition and the iris style transfer pipelines require an eye segmentation model to extract style features from the iris region. For the experiments using the OpenEDS2019 dataset, we applied the pre-trained RITnet model~\cite{chaudhary2019ritnet} due to its excellent performance in real-time precise eye region segmentation. As for the OpenEDS2020 dataset, since the two datasets are different in many aspects, especially image resolution, RITnet does not perform well on OpenEDS2020~\cite{feng2022real}. Therefore we switched to a pre-trained EfficientNet~\cite{openeds2020_seg_model} for eye semantic segmentation for the OpenEDS2020 dataset, which was the winner of OpenEDS2020 segmentation challenge~\cite{openeds2020_seg_comp}. 
\setlength{\myheightfirst}{1.8cm}
\begin{figure}[!ht]
    \centering
    \subfloat[raw]{\includegraphics[height=\myheightfirst, keepaspectratio]{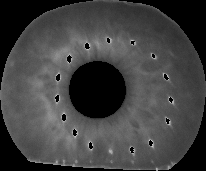} 
    \label{fig:iris}}
    \subfloat[$relu1\_1$]{\includegraphics[height=\myheightfirst, keepaspectratio]{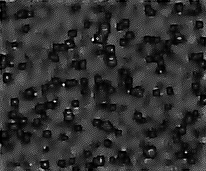}
    \label{fig:relu1_1}}
    \subfloat[$relu2\_1$]{\includegraphics[height=\myheightfirst, keepaspectratio]{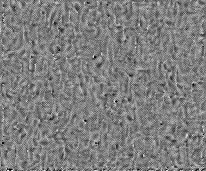}
    \label{fig:relu2_1}}
    \subfloat[$relu3\_1$]{\includegraphics[height=\myheightfirst, keepaspectratio]{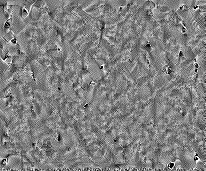} 
    \label{fig:relu3_1}}
    \subfloat[$relu4\_1$]{\includegraphics[height=\myheightfirst, keepaspectratio]{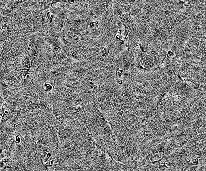}
    \label{fig:relu4_1}}
    \subfloat[$relu5\_1$]{\includegraphics[height=\myheightfirst, keepaspectratio]{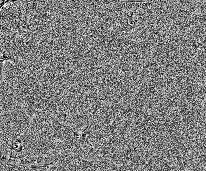}
    \label{fig:relu5_1}}
    \caption{Style reconstructions using a single layer each time. The raw iris sources from OpenEDS2019~\cite{garbin2019openeds} dataset.}
    \label{fig:reconstructions}
    \Description{Style reconstructions using a single layer each time.}
\end{figure}

We employed a pre-trained VGG19~\cite{simonyan2014very} for feature extraction, as it remains a gold standard for neural style transfer due to its effectiveness in capturing detailed style information. Since VGG19 requires colorful images as input, we duplicated the grayscale iris textures over the color channel, and resized them accordingly. To determine which layers of the VGG19 network shall be used for style feature extraction, we computed the style reconstructions using a single layer each time by setting the content objective weight $\alpha$ to 0 and initializing with random noise, and compared the obtained reconstructions in Figure~\ref{fig:reconstructions}. As most of the style information can be captured with the first activation layers of the first four convolutional blocks ($relu1\_1, relu2\_1, relu3\_1, relu4\_1$), we fixed them as the feature extracting layers for our prototype with equal weights. In the end, the batch normalization statistics of these layers were concatenated together, resulting in a 1D vector of length 1,920. The layer $relu4\_2$ was adopted for content loss calculation. In contrast to the Adam optimizer used for other tasks, the LBFGS optimizer with a learning rate of 1.0 was adopted for iris style transfer, as suggested in~\cite{lbfgs}.

We benchmarked the iris recognition performance using iris style features against the conventional feed-forward CNN features, which can be obtained at the last encoding layer of the VGG19 network. For each feature set, we used a classification head that resembles the architecture of the original VGG projection head as much as possible. When training the projection heads, we fixed the batch size to 64 and used the Adam optimizer for classifier training. We conducted a grid search for the optimal learning rate beforehand. The overall best-performing learning rate was 1e-5 for both heads. Once trained, the recognition models were frozen in the subsequent experiments.

To explore the influence of iris style transfer on gaze estimation, we employed both model- and appearance-based gaze estimation models. The former first sends eye images through an eye segmentation model, which is EfficientNet in our case, and then fits ellipses to the pupil, iris, and sclera regions. In the next step, 19 eye landmark features, such as pupil center, pupil angles, and eye corners, are extracted and fed into a regression head in the end. The latter harnesses a pre-trained ResNet50~\cite{he2016deep} as a feature extractor, coupled with a projection head that resembles the previous one. The model-based gaze estimator was simplified from DeepVOG~\cite{yiu2019deepvog, feng2022real}, while the appearance-based model was suggested in~\cite{palmero2020openeds2020}. When training the two gaze estimators, we applied a batch size of 128 and an Adam optimizer with an optimal learning rate of 1e-5. Similar to the iris recognizers, after being trained, the gaze estimators were frozen for downstream benchmarks.

\subsection{Iris Style Feature-Driven User Recognition} \label{sec:recog}
\subsubsection{Feasibility of Style-Driven Recognition} \label{sec:feasibility}
\begin{figure}[!ht]
    \centering
    \subfloat{\includegraphics[height=3.5cm, keepaspectratio]{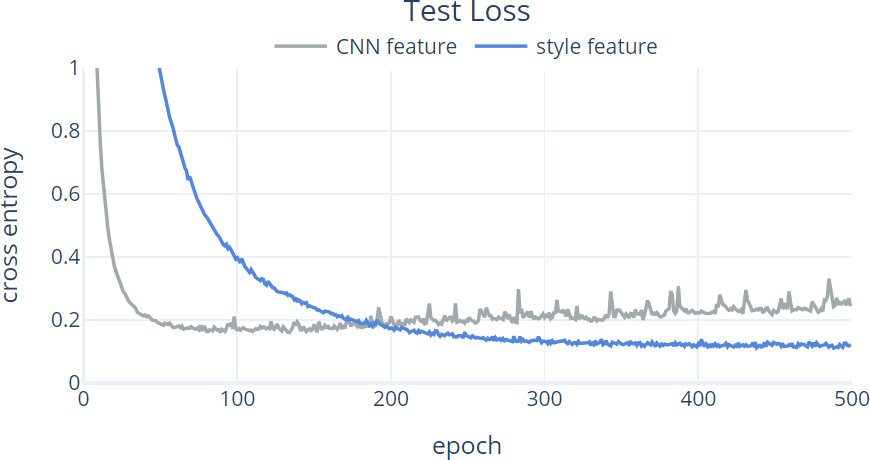} 
    \label{fig:loss}}
    \quad
    \subfloat{\includegraphics[height=3.5cm, keepaspectratio]{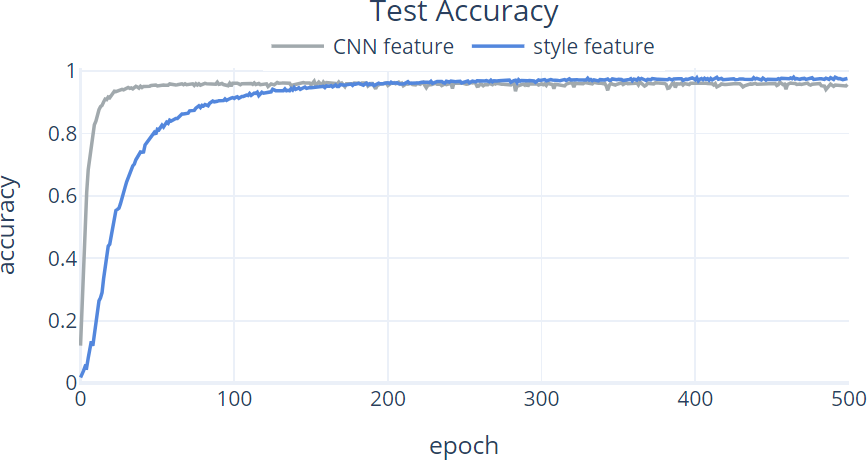}
    \label{fig:accuracy}}
    \caption{Test performance of style feature and common CNN feature for iris recognition.}
    \label{fig:feasibility}
    \Description{Test performance of style feature and common CNN feature for iris recognition.}
\end{figure}

We first investigated the feasibility of iris style feature-based user recognition. We trained both classification models for 500 epochs without image variation and plotted the achieved test loss and accuracy in Figure~\ref{fig:feasibility}. Other metrics, including F1 score and Matthews correlation coefficient (MCC), are reported in the appendix. We observed that while the common CNN feature converged faster than the style feature, it started to overfit after roughly 70 epochs, as the test loss began to increase, and only reached a test accuracy of 95.7\% in the end. In contrast, although the style feature-based classifier converged relatively slower, it never suffered from overfitting and achieved an accuracy of 97.6\% in convergence. Such experiment results endorse the viability of the iris style feature for iris recognition.

\subsubsection{Robustness against Random Rotation and Perspective Transformation} \label{sec:robustness}
\begin{figure}[!ht]
    \centering
    \subfloat[raw]{\includegraphics[height=2.2cm, keepaspectratio]{figures/iris.png} 
    \label{fig:raw}}
    \quad
    \subfloat[rotated]{\includegraphics[height=2.2cm, keepaspectratio]{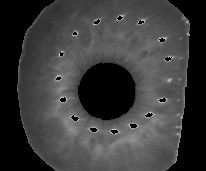}
    \label{fig:rotated}}
    \quad
    \subfloat[perspective shifted]{\includegraphics[height=2.2cm, keepaspectratio]{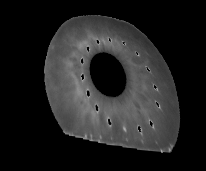}
    \label{fig:perspected}}
    \caption{Examples of image variations. The raw iris sources from OpenEDS2019~\cite{garbin2019openeds} dataset.}
    \label{fig:variation}
    \Description{Examples of image variations.}
\end{figure}

We further introduced image variations in the form of random rotation and random perspective shift of different degrees into the dataset and benchmarked the resilience of both feature sets to the variations. Both image variations are common problems of camera-based sensor systems. We gave two examples of the variations in Figure~\ref{fig:variation} for better understanding. To serve the task, we took the classifiers that were trained in the last task and froze them from gradient updates. For image rotation, we repeated the experiments, iterating over rotation degrees in the range $\{5, 10, 20, 30, 45, 60, 90, 120, 150, 180\}$. Given a rotation degree $d$, each iris texture was rotated randomly by an angle from $(-d, +d)$. For perspective transformation, the experiments were also duplicated multiple times, where every image was shifted by a degree in the range $\{0.01, 0.05, 0.1, 0.2, 0.3, 0.4, 0.5, 0.6, 0.7, 0.8, 0.9, 1.0\}$. 

\begin{figure}[!t]
    \centering
    \subfloat{\includegraphics[height=3.5cm, keepaspectratio]{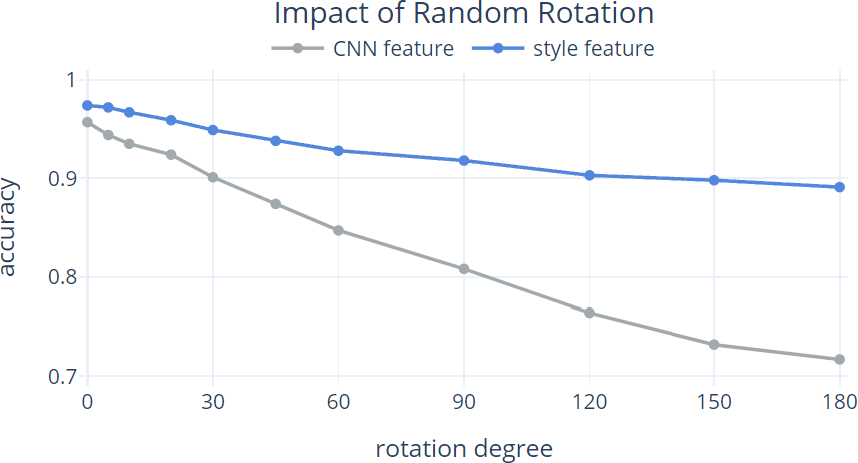} 
    \label{fig:rotation_accu}}
    \quad
    \subfloat{\includegraphics[height=3.5cm, keepaspectratio]{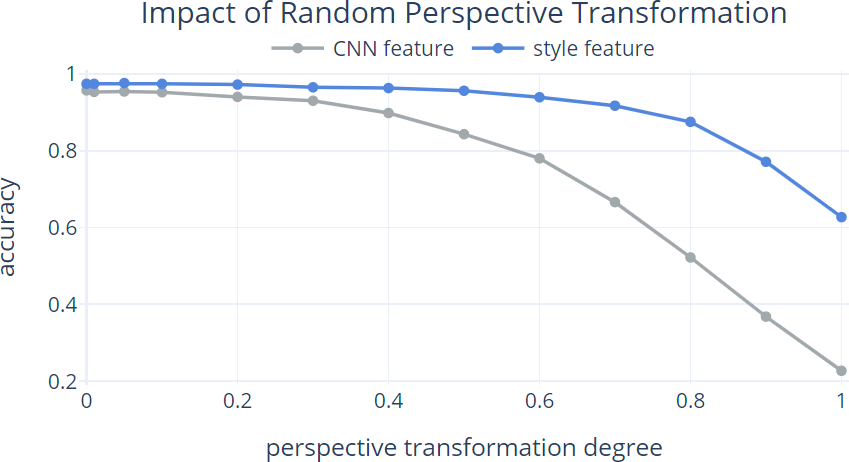}
    \label{fig:perspective_accu}}
    \caption{Iris style feature vs. common CNN feature regarding robustness against random rotation and perspective transformation.}
    \label{fig:variation_accu}
    \Description{Iris style feature vs. common CNN feature regarding robustness against random rotation and perspective transformation.}
\end{figure}

The performance of both feature sets, measured by recognition accuracy on the test set, was recorded in Figure~\ref{fig:variation_accu}. We noticed a close to linear decrease in test accuracy for both feature sets with the increase in rotation degree, while the style feature, with a substantially lower decreasing slope, outperformed the conventional CNN feature to a large extent. For perspective transformation, both feature sets were rather resilient to distortion at low shift levels, with the style feature being in the lead, whereas after a certain breakpoint (distortion degree of around 0.4), the test accuracy began to drop noticeably, and the style feature kept its leading position. We clearly perceived that for random rotation and perspective transformation, the iris style feature-based classifier surpassed the common CNN classification model significantly across different distortion degrees, which validates our claim about the robustness of the iris style feature.

\subsection{Influence of Iris Style Transfer} \label{sec:impact}
\subsubsection{Impact on Image Quality} \label{sec:quality}
Having demonstrated the feasibility and robustness of the iris style feature for recognition purposes, we forwarded our attention to privacy preservation via iris style transfer. 
\setlength{\myheightsecond}{1.7cm}
\begin{table}[!h]
    \centering
    \caption{Examples of stylized iris ($\alpha = \beta = 1.0$, epoch = 200). The content and style irises source from OpenEDS2019~\cite{garbin2019openeds} dataset.}
    \label{tab:iris_nst}
    \begin{tabular}{c|ccc}
         \includegraphics[height=\myheightsecond, width=\myheightsecond]{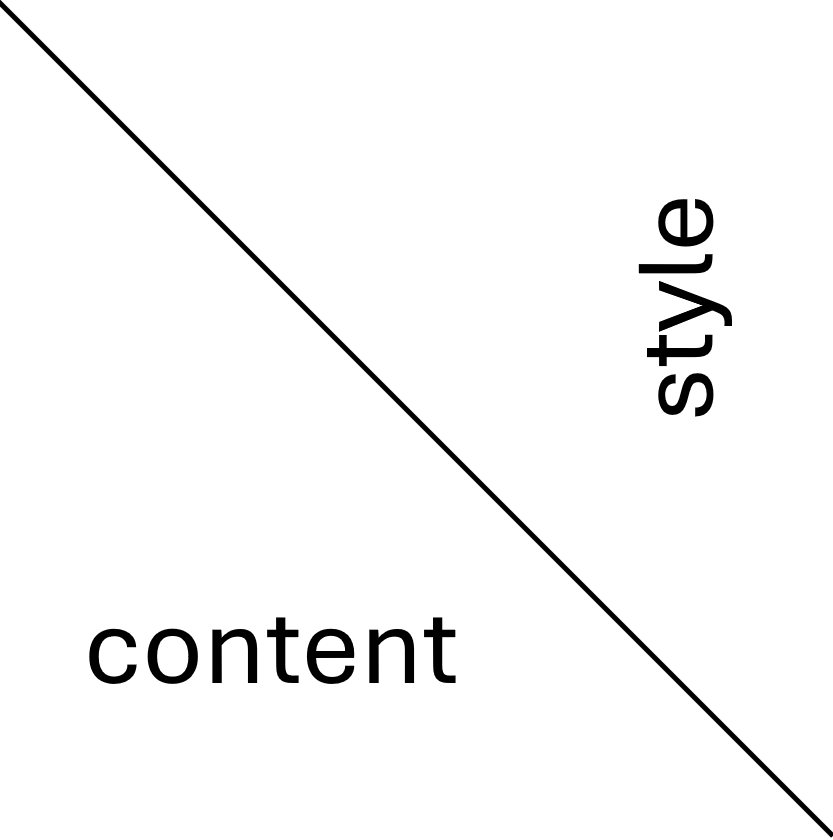} & \includegraphics[height=\myheightsecond]{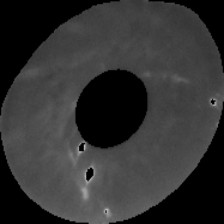} & \includegraphics[height=\myheightsecond]{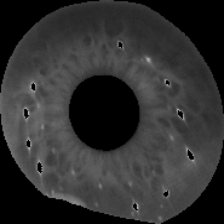} & \includegraphics[height=\myheightsecond]{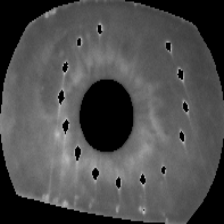}  \\
        \hline
        \\ [-0.7em]
        \includegraphics[height=\myheightsecond]{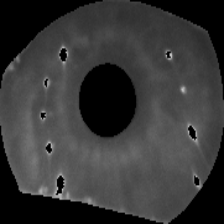} & \includegraphics[height=\myheightsecond]{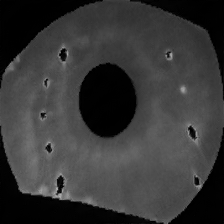} & \includegraphics[height=\myheightsecond]{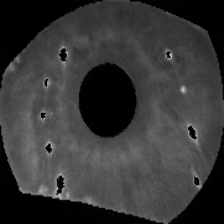} & \includegraphics[height=\myheightsecond]{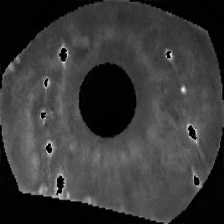} \\
        \\ [-0.7em]
        \includegraphics[height=\myheightsecond]{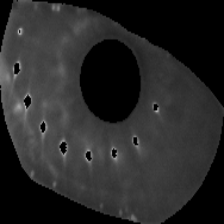} & \includegraphics[height=\myheightsecond]{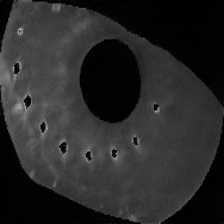} & \includegraphics[height=\myheightsecond]{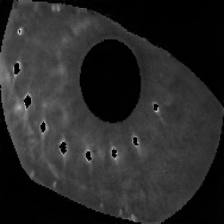} & \includegraphics[height=\myheightsecond]{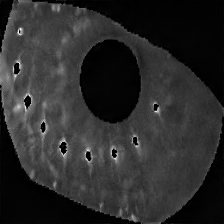} \\
        \\ [-0.7em]
        \includegraphics[height=\myheightsecond]{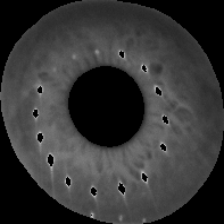} & \includegraphics[height=\myheightsecond]{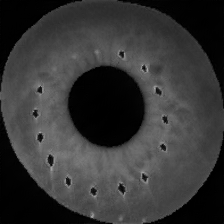} & \includegraphics[height=\myheightsecond]{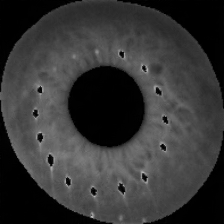} & \includegraphics[height=\myheightsecond]{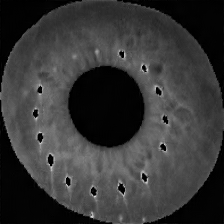}
    \end{tabular}
\end{table}
To this end, we loaded the classification models trained in the last tasks and froze them when investigating iris style transfer. In the next tasks, we mainly focused on the impact of iris style transfer on image quality, privacy, image utility, and risk of malicious use. We first had an insight into the image quality and realism of stylized iris regions. Some of the stylization outputs were listed in Table~\ref{tab:iris_nst}. As depicted, the recombination of content and style has changed iris patterns regarding their visual representations to a certain degree, while the stylized iris regions still looked rather realistic, with image quality on par with their originals, compared to previous works like~\cite{chaudhary2020privacy, narkar2024swap}. Moreover, the synthesized images generally resembled the original content images. Such observations indicate that the iris style transfer pipeline can maintain image quality to a large extent.

\subsubsection{Impact on Privacy} \label{sec:privacy}
As iris style transfer preserves iris quality and realism, we moved our concern to the privacy-preservation effect of our pipeline, which is measured by its influence on recognition performance. Given an eye image of user $A$, we randomly sampled an eye image of another user $B$ and then transferred the iris style of $B$'s image into $A$'s iris texture, and compared the recognition accuracy before and after style transfer. As the stylization effect depends on the ratio between loss weights $\alpha$ and $\beta$, we fixed the weight for content objective $\alpha$ to 1 and then investigated the influence of style loss weights $\beta$ on recognition performance, with the choice of $\beta$ covering $\{1e-4, 1e-3, 1e-2, 1e-1, 1e0, 1e1, 1e2, 1e3, 1e4\}$. Another impacting factor is the number of transfer iterations, for which we tried different values from $\{1, 5, 10, 20, 50, 100, 150, 200\}$.
\begin{figure}[!ht]
    \centering
    \subfloat{\includegraphics[height=5.2cm, keepaspectratio]{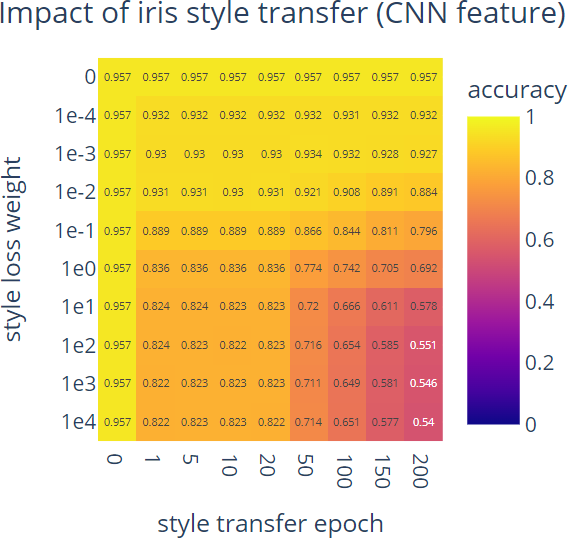} 
    \label{fig:heatmap_cnn}}
    \qquad
    \subfloat{\includegraphics[height=5.2cm, keepaspectratio]{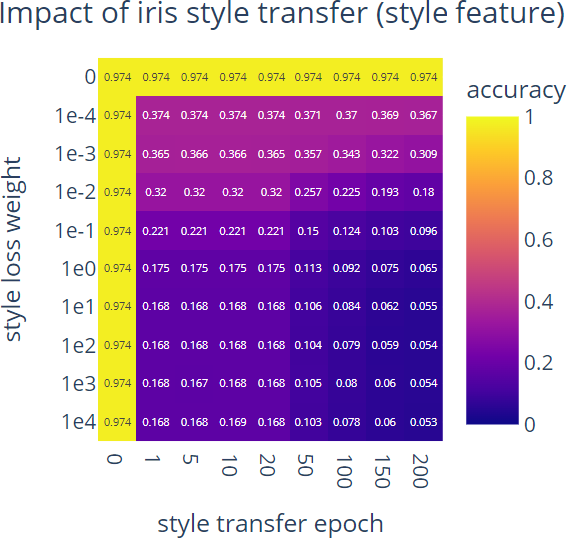}
    \label{fig:featmap_style}}
    \caption{Influence of style loss weight ($\beta$) and style transfer epoch on iris recognition for both feature sets.}
    \label{fig:heatmaps}
    \Description{Influence of style loss weight ($\beta$) and style transfer epoch on iris recognition for both feature sets.}
\end{figure}

The influence of both factors was depicted in Figure~\ref{fig:heatmaps}. In the heatmaps, we saw a substantial drop in recognition accuracy using style features when the iris style was transferred, even in case the transfer was executed solely for a single epoch with tiny $\beta$. In detail, when transferring the iris style for one epoch with a subtle style objective weight $\beta$ of $1e-4$, the test accuracy reduced notably from 97.4\% to 37.4\%. With the heaviest transfer setting, i.e., transferring for 200 epochs and $\beta = 1e4$, the classification accuracy shrinked to 5.3\%. The results show that iris style transfer can effectively block style feature-based iris recognition. In comparison, although iris style transfer did not directly influence the conventional CNN feed-forward feature, it still lowered the recognition performance of the CNN feature-based classifier to a large extent: with a transfer of 200 iterations and $\beta = 1e4$, the test accuracy of CNN feature-driven iris classification fell from 95.7\% to 54.0\%. 

\subsubsection{Impact on Image Utility - Eye Segmentation} \label{sec:segmentation}
Moreover, we investigated the impact of iris style transfer on the utility of eye images. In this regard, we first studied the influence of our pipeline for eye region segmentation. We fed the stylized eye images into RITnet again and compared the achieved per-class Intersection over Union (IoU) metrics before and after stylization. There were four segmentation classes in total, including skin and background (class 0), sclera (class 1), iris (class 2), and pupil (class 3). We set the style loss weight to $\beta = 1.0$ and ran the transfer for 200 epochs. We list the obtained results in Table~\ref{tab:ious}. A closer look at the impact of iris style transfer on IoU distribution across the dataset was visualized in Figure~\ref{fig:ious} in the appendix. As shown in the table and the figure, iris style transfer introduced negligible to zero drops in IoU for each class. The segmentation accuracy for the background remained unchanged, whereas the IoU for the sclera and iris regions dropped slightly from 0.932 to 0.925 and from 0.956 to 0.945, respectively. The mean IoU over class decreased merely from 95.7\% to 94.7\% on average. These results indicate a minor impact on eye image utility from the segmentation perspective. 
\begin{table}[!ht]
    \caption{Influence of iris style transfer on eye segmentation accuracy (averaged over test set).}
    \label{tab:ious}
    \begin{tabular}{lccccc}
     & IoU (skin) & IoU (sclera) & IoU (iris) & IoU (pupil) & mIoU \\
     \hline
     pre-transfer & 0.995 & 0.932 & 0.956 & 0.945 & 0.957 \\
     \hline
     post-transfer & 0.995 & 0.925 & 0.945 & 0.924 & 0.947
    \end{tabular}
\end{table}

\subsubsection{Impact on Image Utility - Gaze Estimation} \label{sec:estimation}
Further, we shifted our focus to the influence of iris style transfer on gaze estimation. For this purpose, we trained both model- and appearance-based gaze estimation models on the OpenEDS2020 dataset. The former yielded a 6.97° gaze error on average on the test set, whereas the latter achieved an error of 3.19°. They respectively matched the baseline and the winner performance of OpenEDS2020 gaze prediction challenge~\cite{openeds2020_est_comp}. For the design of both models, we referred readers to subsection~\ref{sec:models}.
\begin{table}[!h]
    \caption{Influence of iris style transfer on gaze estimation error (averaged over test set).}
    \label{tab:gaze_errors}
    \begin{tabular}{lcc}
     & gaze error (model-based) & gaze error (app-based) \\
     \hline
     pre-transfer & 6.97° & 3.19° \\
     \hline
     post-transfer & 7.02° & 3.25°
    \end{tabular}
\end{table}

After we trained the gaze estimators, we froze them from gradient updates and compared their performance before and after conducting iris style transfer. Similar to the segmentation task, we fixed $\beta = 1.0$ and ran the transfer for 200 epochs. The obtained gaze estimation error was given in Table~\ref{tab:gaze_errors}. Additionally, we visualized the gaze error distributions before and after style transfer in Figure~\ref{fig:gaze_error} in the appendix. As depicted in the table, the gaze estimation errors in degree increased by 0.05° for the model-based estimator and by 0.06° for the appearance-based model. Both increases were negligible compared to the gaze errors themselves, which shows that our iris style transfer pipeline maintains the utility of eye images regarding eye tracking.

\subsubsection{Risk of Malicious Use} \label{sec:risk}
Furthermore, we were interested in the false acceptance rate (FAR) caused by iris style transfer, i.e., the probability of misidentifying user $A$ as user $B$ when $B$'s iris style is transferred into $A$'s iris texture. FAR is commonly computed as the ratio between the number of wrongly accepted users and the total number of imposter attempts. This metric is important because it implies the risk that the system incorrectly accepts an unauthorized user, which is, in our case, the possibility of malicious use of iris style transfer to fake someone's presentation, similar to what has already been realized in~\cite{narkar2024swap} with the underlying technique from~\cite{chaudhary2020privacy}. In our experiments, for each eye image of user A, we randomly sampled another eye image of another user B and transferred the iris style feature from B's image into A's iris region. In other words, the total number of imposter attempts in our experiments was equal to the total number of samples in the dataset. Similar to previous experiments, we set $\beta = 1.0$ and ran the transfer for 200 epochs. We list our results in Table~\ref{tab:far}. Based on the results, iris style transfer slightly escalated the FAR to 1.1\% for the CNN feature-driven classifier and to 6.1\% for the style feature-based classification model. As both FARs are still low, the risk of misuse remains relatively low, in contrast to previous works~\cite{chaudhary2020privacy, narkar2024swap}, where the whole iris texture was replaced, and the potential FAR can be quite high. 
\begin{table}[!ht]
    \caption{Risk of malicious use of iris style transfer, measured by FAR (false acceptance rate).}
    \label{tab:far}
    \begin{tabular}{lcc}
     & FAR (CNN feature) & FAR (style feature) \\
     \hline
     pre-transfer & 0.001 & 0.000 \\
     \hline
     post-transfer & 0.011 & 0.061
    \end{tabular}
\end{table}

\section{Discussion}
We proposed a novel framework for iris recognition by introducing neural style transfer as a feature extraction method. Through extensive experimentation, we demonstrate that iris style features, represented as statistical distributions rather than traditional CNN embeddings, are a valuable feature modality for iris recognition and outperform conventional CNN embeddings on recognition accuracy.
We identified two main reasons for this superiority. Firstly, we suspect that iris recognition is intrinsically more a style-driven process than content-driven, due to the uniqueness and identifiability of style features~\cite{wright2022artfid}. Secondly, style features are of lower dimensionality compared to traditional CNN features, resulting in reduced model complexity and overfitting. A further benefit of iris style features is that their dimensionality is constant regardless of the resolution of input images. For instance, provided that the layers $relu1\_1, relu2\_1, relu3\_1, relu4\_1$ are used for style feature extraction, the style features, i.e., concatenated distribution means and standard deviations, are consistently a vector of length 1,920, irrespective of the size of iris images. In contrast, the conventional VGG embeddings are of length 25,088 given an input image of size $224 \times 224$, and the complexity escalates to 122,880 when the input image is of shape $640 \times 400$. Such consistency in feature dimensionality can be crucial for embedding-based algorithms since the trimmed iris is often of various sizes, even if the camera output resolution remains the same. While the problem can be somewhat alleviated by image scaling, it should be noticed that image resizing is not a lossless process and can introduce undesired artifacts, which may undermine the recognition result. Moreover, iris style features offer enhanced resilience to image variations like rotation and perspective shift. These variations, often unavoidable in camera-based iris recognition systems, typically degrade recognition accuracy. Yet, our results show that style features maintain higher accuracy under these conditions, outperforming conventional CNN-based methods. 

Furthermore, we explored a privacy-preserving mechanism using style transfer to obfuscate identifiable iris patterns. This approach achieves a remarkable reduction in unauthorized recognition accuracy while maintaining the image utility for eye segmentation and gaze estimation, highlighting its applicability for both security and usability. In addition, the risk of impersonation attack can be noticeably lower than prior works~\cite{narkar2024swap, chaudhary2020privacy}. 

While our work validates the feasibility and benefits of style-based iris recognition, our evaluation mainly focuses on deep learning-based tasks using a high-quality collected dataset. Broader evaluations across diverse datasets and real-world conditions, as well as further benchmarks involving conventional recognition metrics such as Hamming distance, are needed to affirm the robustness and generalizability of our work. Future extensions could also explore the impact of iris style transfer on other eye utilities, such as gaze prediction and foveated rendering, ensuring the method’s alignment with the privacy-utility trade-off. Additionally, we do not exclude the potential of transferring non-iris style to iris images, which can also be valuable in special cases. 

From the aspects of practical applications, one direction of future work is to eliminate model redundancy. While existing style transfer methods typically rely on CNNs like VGG and ResNet, which remain as gold standards for style transfer and perform well in practice, they are often tailored for RGB images, whereas iris images are often grayscale. Although this problem can be solved by channel duplicating, the redundancy of model parameters causes unnecessary memory waste and computation burden. A pre-trained CNN model targeting grayscale eye images can be valuable. Further, our proposed style extraction and style transfer pipelines rely on eye segmentation. Without iris texture cropping, styles of undesired image parts like sclera and near-eye skin can be captured, which may undermine the outcome. Two meaningful improvement directions are to loose the dependence on eye segmentation and to develop segmentation models that perform well universally. The sensitivity of eye segmentation models to image variations also needs to be further studied. Another limitation of our work is that it is bound to static eye images and is an image-optimization-based online algorithm~\cite{jing2019neural}, which suffers from high computation overhead due to iterative forward and backward propagations. To this end, we refer to the integration of model-optimization-based offline algorithms that approach style transfer with a simple and fast feed-forward using pre-trained networks~\cite{huang2017real, gao2019reconet, liu2021adaattn} as future work. With the deployment of offline stylization algorithms, our proposed iris style transfer prototype has a good potential to approach real-time consistent iris stylization on modern devices. Then, a typical use case of the improved pipeline can be, for instance, streaming and recording stylized eye video in real-time for head-mounted displays or mobile eye trackers like Varjo XR-4 that possess good computer power and support real-time eye streaming. Furthermore, we address the application of other style transfer methods, such as GAN-based models like Pix2Pix~\cite{isola2017image} and CycleGAN~\cite{zhu2017unpaired}, to approach iris manipulation, as future work.

\section{Conclusion}
In this paper, we introduced a pioneering approach to iris recognition by leveraging neural style transfer to extract iris style features. Our findings underscore that style features, capturing global statistical properties of iris textures, offer notable advantages over traditional CNN features by maintaining higher classification accuracy amidst rotational and perspective variations. This resilience to common camera-induced distortions paves the way for more reliable, adaptable iris recognition systems, particularly in consumer devices like VR/AR head-mounted displays and smart glasses where stable, high-quality images may not always be guaranteed.

We also proposed a privacy-preserving method, utilizing style transfer to obfuscate original iris patterns, which represents a significant advancement in protecting iris-based biometrics. By effectively obscuring identifiable features while preserving the functionality of the obfuscated eye images for tasks like eye segmentation and gaze estimation, our approach introduces a viable solution for balancing biometric security with privacy concerns and data utility. This research opens new directions in biometric systems that prioritize both accuracy and privacy. Moving forward, expanding the application of iris style transfer to dynamic, real-time processing and further refining privacy safeguards will strengthen the potential for deployment across diverse environments and device platforms, making style transfer a reasonable tool for the next generation of secure, privacy-conscious biometric technologies.

\begin{acks}
We acknowledge the funding by the Deutsche Forschungsgemeinschaft (DFG, German Research Foundation) – Project number KA 4539/5-1.
\end{acks}

\bibliographystyle{ACM-Reference-Format}
\bibliography{reference}


\newpage
\appendix
\section*{Appendix} \label{sec:appendix}

\section{Implementation}
We published our code on~\url{https://gitlab.lrz.de/hctl/Iris-Style-Transfer} for reproducibility.

\section{Randomness}
To guarantee reproducibility, we fixed random seed to 42 at the very beginning of our program. The random seed was fed to $numpy.random.seed$, $torch.manual\_seed$, and $random.seed$. 

\section{Datasets}
The OpenEDS2019 dataset~\cite{garbin2019openeds} contains four parts, namely: segmentation data, generative data, sequence data, and corneal topography data. We considered only the segmentation data. The OpenEDS2020 dataset~\cite{palmero2020openeds2020, palmero2021openeds2020} contains two subsets, respectively for gaze estimation and sparse eye segmentation. We used only the gaze estimation set. The details of both datasets are given in Table~\ref{tab:appendix_datasets}.
\begin{table}[!ht]
\caption{Overview of the OpenEDS2019 and OpenEDS2020 datasets.}
\label{tab:appendix_datasets}
\begin{tabular}{l c c}
    \textbf{Property} & \textbf{OpenEDS2019 (segmentation)} & \textbf{OpenEDS2020 (estimation)} \\
    \hline
    samples & 12,759 & 550,400 \\
    users & 152 & 80 \\
    modality & grayscale image & grayscale image \\
    resolution & $400 \times 640$ & $640 \times 400$ \\
    ground truth & segmentation labels & 3D gaze vectors
\end{tabular}
\end{table}

\section{Environment}
Our environment by date February 1st, 2025 was listed in Table~\ref{tab:appendix_environment}. It should be noticed that although our code can be run on a device without dedicated GPU, it is recommended to run the program on a computer with $\geq$ 32GB RAM and a dGPU with $\geq$ 32GB VRAM.
\begin{table}[!ht]
    \caption{Experiment environment.}
    \label{tab:appendix_environment}
    \resizebox{\textwidth}{!}{%
    \begin{tabular}{l c}
        \textbf{Hardware} & \textbf{Specification}  \\
        \hline
        CPU & AMD EPYC 7763 64-Core \\
        GPU & NVIDIA A100 80GB PCIe$\times$4 \\
        Memory & 1TB \\
    \end{tabular}
    \qquad
    \begin{tabular}{l c}
        \textbf{Software} & \textbf{Version} \\
        \hline
        OS & Ubuntu 22.04.5 LTS \\
        Python & 3.12.7 by Anaconda \\
        PyTorch & 2.6.0 for CUDA 12.6 \\
        OpenCV-Python & 4.11.0.86 \\
        Scikit-Learn & 1.6.1 \\
        Scikit-Image & 0.25.1 \\
        segmentation-models-pytorch & 0.4.0 \\
        WandB & 0.19.5 \\
    \end{tabular}
    }
\end{table}

\section{Hyperparameters} 
For training the classifiers, we used the Adam optimizer and conducted grid search in range $\{1e-6, 1e-5, 1e-4, 1e-3, 1e-2\}$ for optimal learning rates. The best performing learning rates were $1e-5$ for both feature sets. For style transfer, we followed the instruction given by~\cite{lbfgs}, namely using an LBFGS optimize with learning 1.0. We used a batch size of 64 for the aforementioned tasks. For training gaze estimators, similar to fitting the classifiers, we applied the Adam optimizer with optimal learning rate $1e-5$ for both estimation models. Batch size of 128 was picked during training. The hyperparameters were listed in Table~\ref{tab:appendix_hyperparameters} for a clear view.
\begin{table}[!ht]
    \caption{Details of hyperparameters.}
    \label{tab:appendix_hyperparameters}
    \centering
    \begin{tabular}{l c}
        \textbf{Hyperparameter} & \textbf{Value} \\
        \hline
        optimizer (classifiers) & Adam \\
        optimizer (style transfer) & LBFGS \\
        optimizer (gaze estimators) & Adam \\
        learning rate (classifiers) & $1e-5$ \\
        learning rate (style transfer) & $1.0$ \\
        learning rate (gaze estimators) & $1e-5$ \\
        batch size (classifiers) & 64 \\
        batch size (style transfer) & 64 \\
        batch size (gaze estimators) & 128 \\
        image scaling & $224 \times 224$
    \end{tabular}
\end{table}

\section{Model Structures}
We implemented two different classification heads. The first one was designed for conventional VGG features, whereas the second was tailor for style features. We designed the structures of both heads so that they resemble the original VGG projection head as much as possible. For the gaze estimators, the projection heads were regression MLPs, which output normalized 3D gaze vectors. For detailed architectures of the models, we refer readers to~\url{https://gitlab.lrz.de/hctl/Iris-Style-Transfer/-/tree/main/models?ref_type=heads}.

\section{Additional Results}
Additional results, such as F1 scores, Matthews correlation coefficients (MCC), and distribution plots, were given here.
\begin{figure}[!ht]
    \centering
    \subfloat{\includegraphics[height=3.5cm, keepaspectratio]{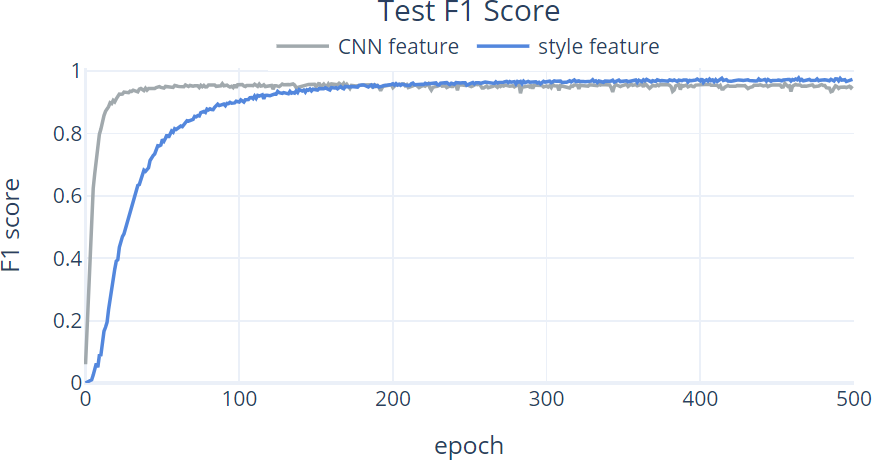} 
    \label{fig:f1}}
    \quad
    \subfloat{\includegraphics[height=3.5cm, keepaspectratio]{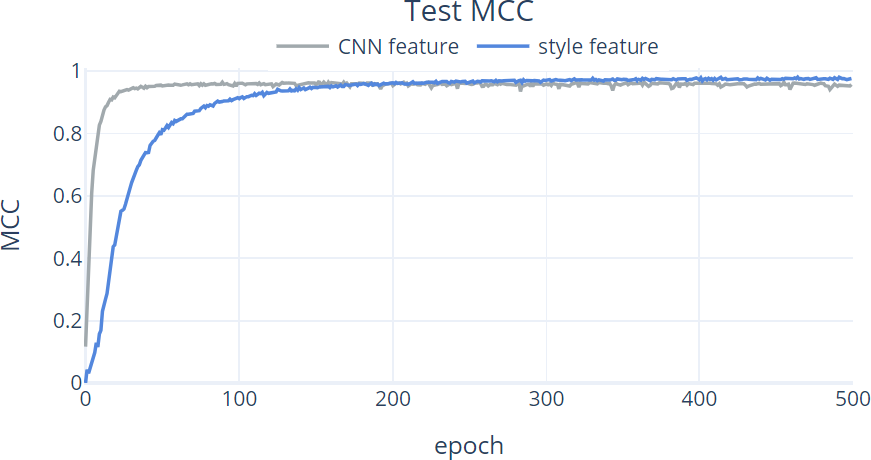}
    \label{fig:mcc}}
    \caption{Test performance of style feature and common CNN feature for iris recognition.}
    \label{fig:appendix feasibility}
    \Description{Test performance of style feature and common CNN feature for iris recognition.}
\end{figure}

\begin{figure}[!ht]
    \centering
    \subfloat{\includegraphics[height=3.5cm, keepaspectratio]{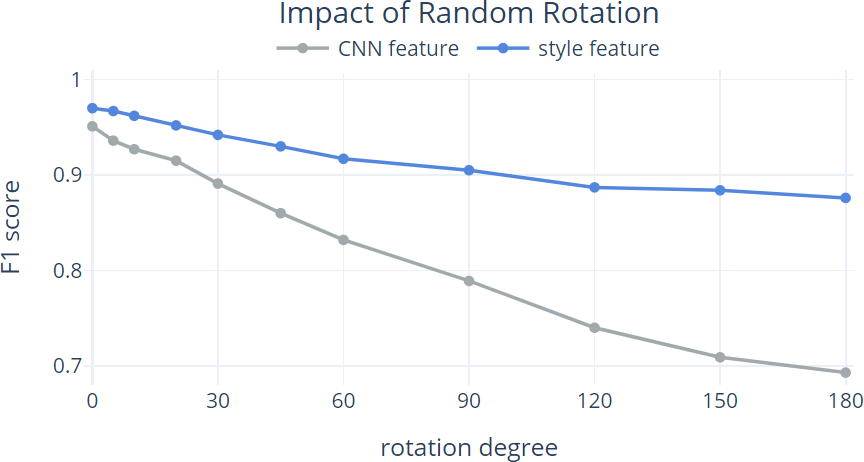} 
    \label{fig:rotation_f1}}
    \quad
    \subfloat{\includegraphics[height=3.5cm, keepaspectratio]{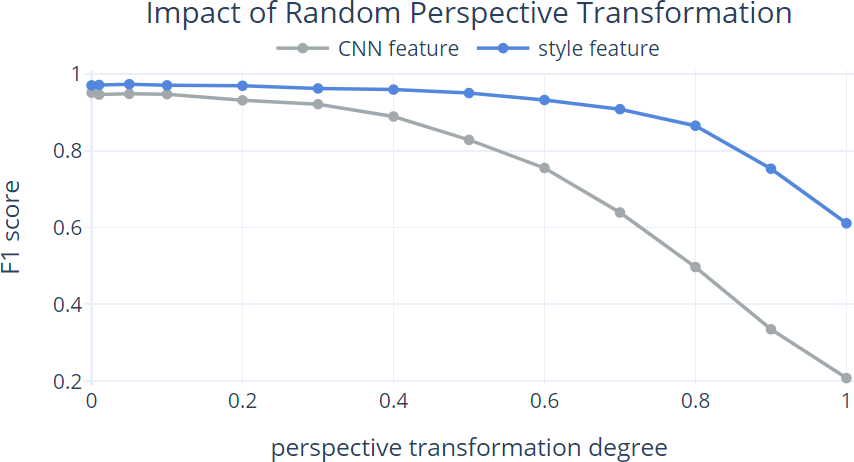}
    \label{fig:perspective_f1}}
    \caption{Style feature vs. common CNN feature regarding robustness against random rotation and perspective transformation (F1 score).}
    \label{fig:variation_f1}
    \Description{Style feature vs. common CNN feature regarding robustness against random rotation and perspective transformation (F1 score).}
\end{figure}

\begin{figure}[!ht]
    \centering
    \subfloat{\includegraphics[height=3.5cm, keepaspectratio]{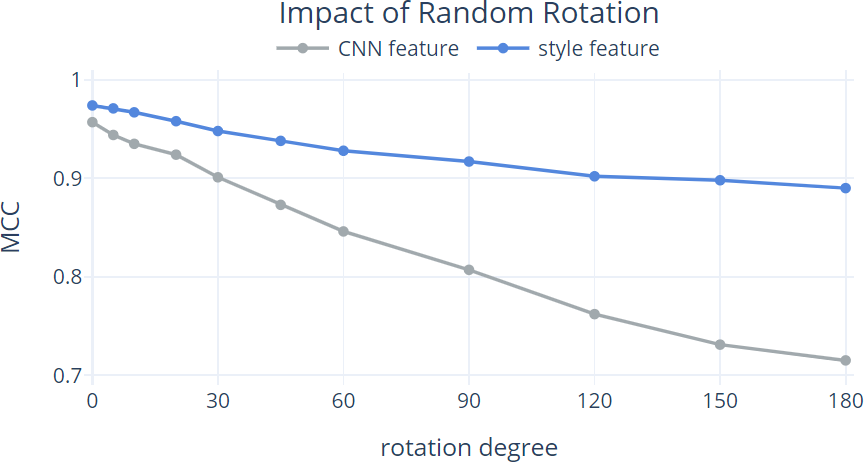} 
    \label{fig:rotation_mcc}}
    \quad
    \subfloat{\includegraphics[height=3.5cm, keepaspectratio]{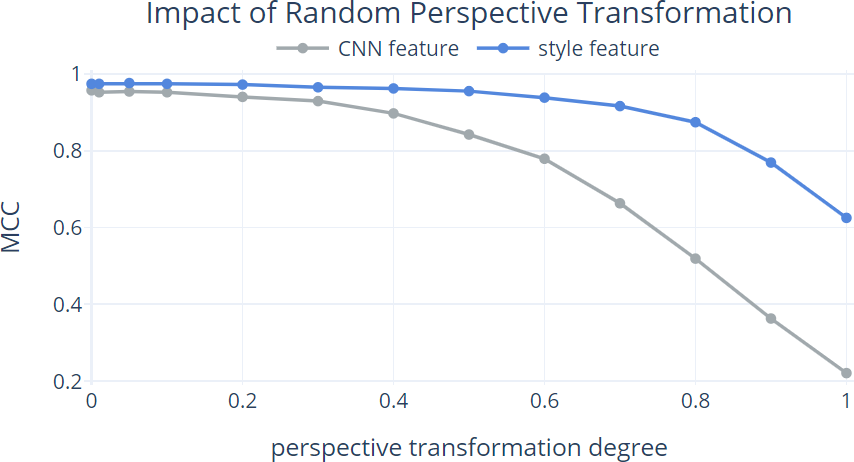}
    \label{fig:perspective_mcc}}
    \caption{Style feature vs. common CNN feature regarding robustness against random rotation and perspective transformation (MCC).}
    \label{fig:variation_mcc}
    \Description{Style feature vs. common CNN feature regarding robustness against random rotation and perspective transformation (MCC).}
\end{figure}

\begin{figure}[!ht]
    \centering
    \subfloat{\includegraphics[height=4.8cm, keepaspectratio]{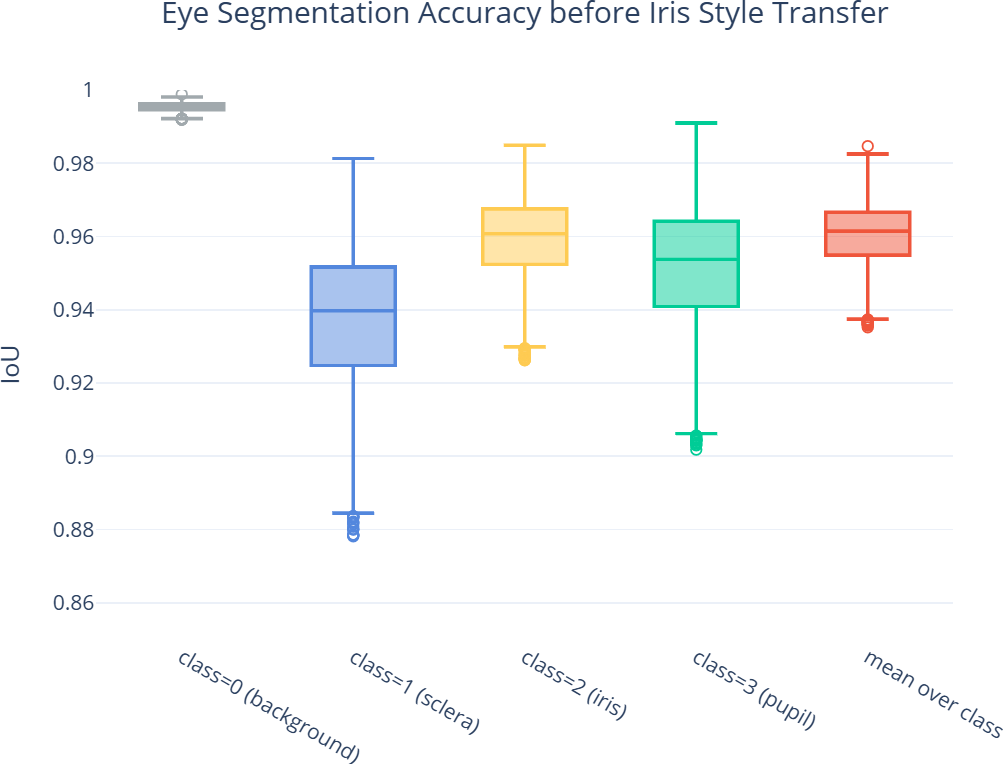} 
    \label{fig:ious_pre}}
    \quad
    \subfloat{\includegraphics[height=4.8cm, keepaspectratio]{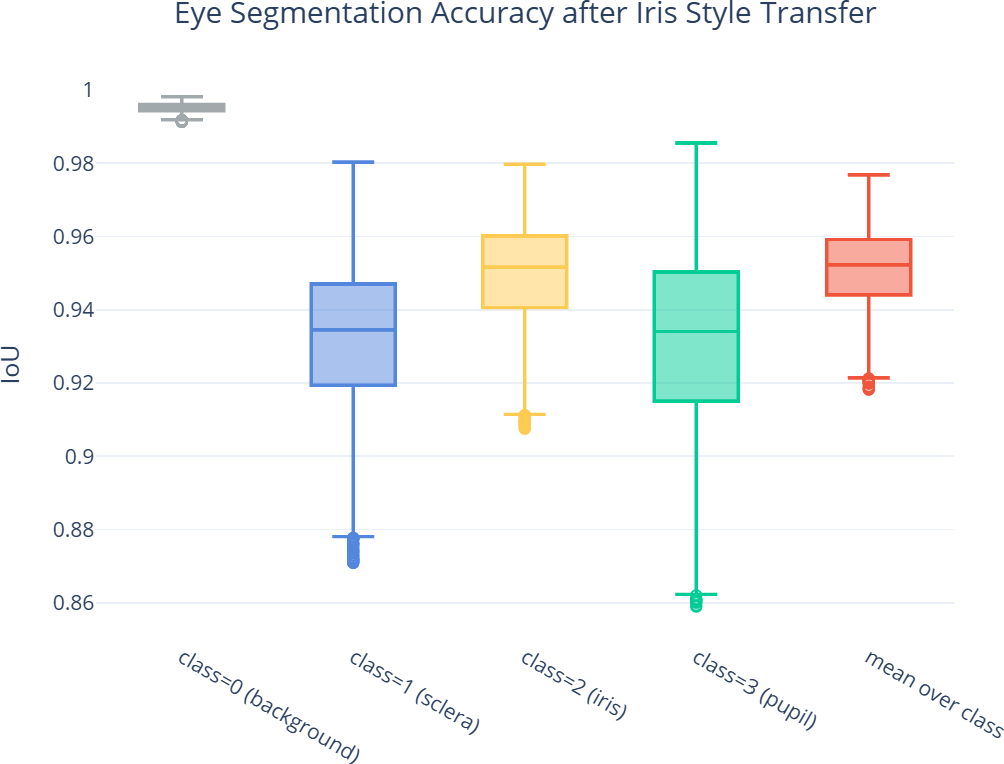}
    \label{fig:ious_post}}
    \caption{Influence of iris style transfer on eye segmentation accuracy (distribution over test set).}
    \label{fig:ious}
    \Description{Influence of iris style transfer on eye segmentation accuracy (distribution over test set).}
\end{figure}

\begin{figure}[!ht]
    \centering
    \subfloat{\includegraphics[height=4.8cm, keepaspectratio]{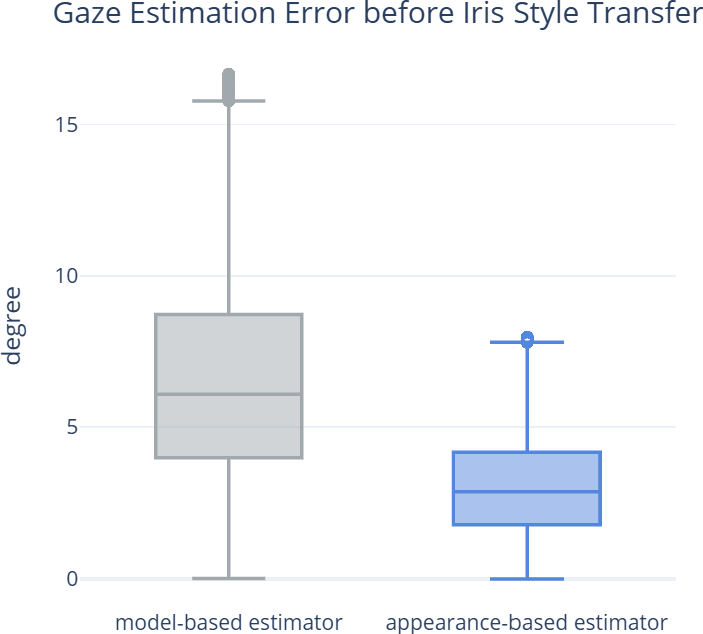} 
    \label{fig:gaze_pre}}
    \quad
    \subfloat{\includegraphics[height=4.8cm, keepaspectratio]{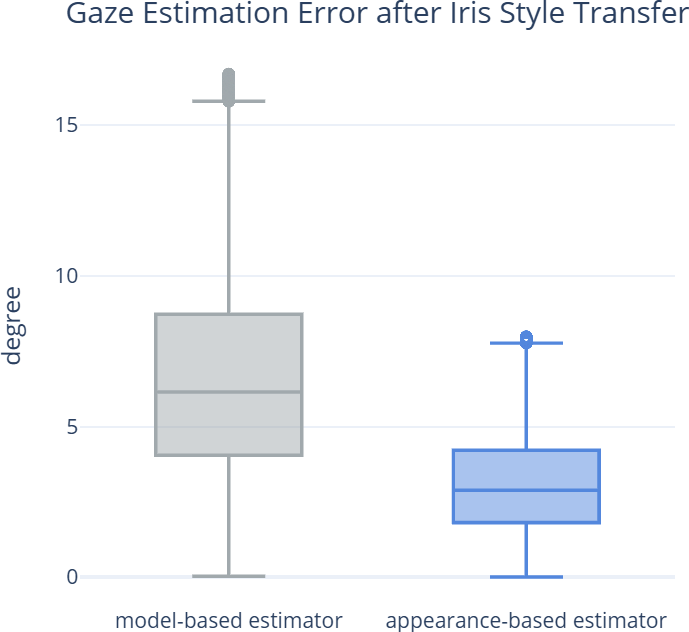}
    \label{fig:gaze_post}}
    \caption{Influence of iris style transfer on gaze estimation error (distribution over test set).}
    \label{fig:gaze_error}
    \Description{Influence of iris style transfer on gaze estimation error (distribution over test set).}
\end{figure}

\end{document}